\documentclass[11pt]{article}

\usepackage[utf8]{inputenc} 
\usepackage[T1]{fontenc}    
\usepackage[margin=1in]{geometry} 
\usepackage[hypertexnames=false,breaklinks=true,draft]{hyperref} 
\usepackage{url}            
\usepackage{natbib}         
\usepackage{booktabs}       
\usepackage{array}          
\usepackage{tabularx}       
\usepackage{amsfonts}       
\usepackage{amssymb}        
\usepackage{amsmath}        
\usepackage{amsthm}         
\usepackage{xcolor}         
\usepackage{graphicx}       
\usepackage{tikz}           
\usetikzlibrary{positioning, fit, backgrounds} 
\usepackage{float}          
\usepackage{mathtools}
\providecommand{\FloatBarrier}{}

\newcounter{researchquestion}
\theoremstyle{definition}
\newtheorem{example}{Example}[section]

\title{Agentic MIP Research: Accelerated Constraint Handler Generation}

\author{%
  Liding Xu\thanks{Equal contribution. Zuse Institute Berlin, Berlin, Germany.
    Correspondence to: \texttt{<lidingxu.ac@gmail.com>}.} \and
  Yugeng Zhou\thanks{Equal contribution. Independent Researcher.
    Correspondence to: \texttt{<yugengzhou@gmail.com>}.} \and
  Sebastian Pokutta\thanks{Zuse Institute Berlin and Technische Universit\"at Berlin, Berlin, Germany.
    Correspondence to: \texttt{<pokutta@zib.de>}.}
}
\date{}

\begin{document}

\maketitle

\begin{abstract}
  Mixed-integer programming (MIP) research is both mathematically sophisticated and engineering-intensive: testing an algorithmic hypothesis within a branch-and-cut solver requires substantial implementation, debugging, tuning, and large-scale benchmarking. We propose an agentic MIP research framework that shortens this feedback loop by embedding LLM agents into a solver-aware harness for generating, verifying, and evaluating plugins for the open-source solver SCIP. Propagation methods play a central role in accelerating MIP solving by exploiting global constraints. We instantiate our framework on the semantic lifting of MIP formulations into global constraints and the automatic construction of propagation-only SCIP constraint handlers. On the MIPLIB 2017 benchmark set, the framework successfully recovers global constraint structures from constraint programming and generates executable constraint detectors and propagation-only constraint handlers. Furthermore, the framework naturally extends to in-context learning within a sandboxed environment, enabling agents not only to tune and debug generated constraint handlers on real instances, but also to explore global constraint patterns in MIP problems and discover novel propagation strategies not yet implemented in SCIP. This framework allows us to systematically distinguish meaningful algorithmic improvements from low-value or overly costly candidates: the novel propagation methods successfully solved five additional instances within the explored benchmark. Overall, this framework demonstrates that LLM agents can autonomously navigate the complex MIP research loop, paving the way for a more automated solver development process.
\end{abstract}

\section{Introduction}
\label{sec:intro}

Mixed-Integer Programming (MIP) is a foundational tool with broad applicability across transportation, energy systems, and machine learning. Beyond core mathematical concepts, the performance of general-purpose MIP solvers relies heavily on implementations of complex and coupled plugins.  While the mathematical principle of an algorithmic idea can often be stated
compactly, hypothesis testing---validating its usefulness within a full solver---remains an engineering-heavy bottleneck. Consequently, improving general-purpose MIP solvers is notoriously difficult. For context, the transition from version 9.0 to 10.0 of the open-source solver SCIP
\citep{achterberg2007,hojny2025scip10} yielded two additional solved instances on the MIPLIB~2017 benchmark~\citep{mip2017}. Similarly, ablation studies on a commercial solver
\citep{achterberg2020presolve} indicate that even after extensive algorithm mining for structural patterns, a successful new plugin typically improves overall solver performance by a
modest 3-5\%.

Generative AI and large language models (LLMs) have shown promise in accelerating solver-independent  modeling tasks (aka translation of natural language to mathematical formulations) \citep{ahmaditeshnizi2023optimus,ahmaditeshnizi2024optimus,ahmed2025chorus,lu2025optmath,jiangllmopt,zhang2024solving}, but these front-end translation approaches  do not directly address the back-end solver-level integration.
A plausible algorithmic hypothesis must be translated into solver plugin  and evaluated on sufficiently large benchmark. Many attractive ideas fail inside a real solver because they duplicate existing mechanisms, interact poorly with numerics, or add overhead that dominates their benefit.  In this view, mixed or negative feedback signals are valuable: they identify not only true-positive ideas but also false-positive or low-value ideas. \textsc{AlphaEvolve}~\citep{novikov2025alphaevolve} demonstrates that in-context learning can identify problem patterns and match them to tailored optimization algorithms (should not be confused with overfitting in supervised learning). Similar frameworks~\citep{sun2024autosat,sun2025automodsat} have leveraged LLM agents to improve SAT solvers, motivating the extension of this approach to more complex MIP solvers.

We argue that LLM agents can be highly effective for MIP research, specifically in generating, verifying, tuning, and discovering solver plugins. Rather than relying on unconstrained
code synthesis, we embed LLM agents within a solver-aware harness designed to produce robust, SCIP-compatible code artifacts. We present an abstract agentic MIP research (agentic
MIPR) framework organized around the following pipeline:
\[
  \text{prompt} \to \text{plugin code generation} \to \text{in-context learning} \to \text{evaluation}.
\]

Our harness engineering grounds the agent in solver APIs, domain-specific knowledge, codebases, and established templates, and it constrains code generation through
well-defined stage contracts. By solving real-world instances, it leverages in-context learning to generate rich feedback signals, enabling algorithmic discovery
beyond standard modeling tasks. The framework provides two task modes.
In few-shot mode, the
agent implements solver plugins for known target families using structured skills and references. In
zero-shot mode, it explores benchmark data for previously unspecified structures and proposes new
plugin candidates.

Many mathematical optimization problems can be modeled in either Constraint Programming (CP) or MIP. Consider a problem parameterized by $c \in \mathcal{C}$ with variables $\mathbf{x}$ and feasible set $\mathcal{X}(c)$. Global constraints
provide a compact semantic description of $\mathcal{X}(c)$. While CP offers a formal  representation of combinatorial structures, MIP modeling
\emph{compiles} them into collections of linear constraints (represented as rows in matrix form) over the original variables $\mathbf{x}$ and auxiliary variables $\mathbf{y}$:
\begin{equation}
  \label{eq:global-linear-map}
  \text{Language}
  \xrightarrow{\text{model}}
  \text{Global Constraint}(\mathbf{x};c)
  \;\xleftrightarrow{\text{encode/decode}}\;
  \text{Linear Constraints}(\mathbf{x},\mathbf{y};c).
\end{equation}
A common principle across both CP and MIP is \emph{primal reduction}: reducing the search space without excluding feasible solutions. Let $D(c)$ be an initial domain box such that
the feasible set satisfies $\mathcal{X}(c) \subseteq D(c)$. A primal reduction rule derives a refined \emph{valid} domain $D'(c)$ satisfying:
\begin{equation}
  \label{eq:valid-reduction}
  \mathcal{X}(c) \subseteq D'(c) \subseteq D(c).
\end{equation}
In CP \citep{regin1994filtering}, such reductions are typically achieved through domain propagation. In MIP solvers, related strengthening mechanisms manifest as either propagation or
cut separation. Because domain reduction is computationally lightweight compared to cut separation, this paper focuses on bound-tightening mechanisms implemented via propagation-only
constraint handlers in SCIP. Then, two questions arise:

\refstepcounter{researchquestion}\label{q:detection}
\noindent\textbf{Q\theresearchquestion.}
\emph{To what extent do linear constraints in MIP formulations encode high-level global constraint semantics, and how can these structures be systematically detected?}

\refstepcounter{researchquestion}\label{q:reduction}
\noindent\textbf{Q\theresearchquestion.}
\emph{Which propagation methods can be leveraged to improve the performance of MIP solvers?}

Rather than relying on  manual research effort, the agentic MIPR framework paves the way to automatically investigate these
questions. The contributions of this paper are as follows.

\begin{itemize}
  \item \textbf{Agentic solver customization:} Automates and accelerates MIP research through hypothesis generation, plugin implementation, and large-scale evaluation.
  \item \textbf{Constraint handler generation as vibe coding:} Generates detectors that parse MIP formulations into CP global constraints, along with corresponding propagation-only constraint handlers.
  \item \textbf{Synthetic verification via reverse sampling:} Verifies the soundness of constraint detectors and propagators on LLM-synthesized MIP instances.
  \item \textbf{Feedback and in-context learning:} Leverages solver execution on real instances and rich solver signals to enable iterative debugging, tuning, and discovery of new constraint patterns and propagation strategies.
  \item \textbf{Improved SCIP performance:} Demonstrates on MIPLIB~2017 the recovery of global constraint structures and the discovery of novel ones, with generated propagation methods solving five additional instances.
\end{itemize}

\section{Background and Related Work}
\label{sec:related-work}

\paragraph{Constraint programming and mixed-integer programming.}
Constraint programming (CP) studies global constraints such as \textsc{AllDifferent}, \textsc{Cumulative}, \textsc{Channel}, and \textsc{NValue}, along with specialized propagation algorithms for pruning infeasible domain values \citep{aggoun1993extending,beldiceanu2005global,regin1994filtering,rossi2006handbook}. When these structures are compiled into MIP formulations, their semantics may be distributed across many linear rows and auxiliary variables. Propagation is called node preprocessing in MIP solvers \citep{achterberg2008constraint}.  Google OR-Tools \citep{ortools} provides a hybrid MIP/CP modeling and solving alternative. Many MIP algoritms can be realized as solver plugins: propagation, cut separation
\citep{tang2020reinforcement}, branching rule \citep{gupta2020hybrid,khalil2016learning}, heuristics, and plugin scheduling \citep{chmiela2021learning}. SCIP \cite{achterberg2007} natively supports customized constraint handlers and propagators. This extensibility creates opportunities to leverage these
mechanisms for diverse applications, such as propagation-based  neural network verification~\citep{wang2018formal}.

\paragraph{LLM-generated optimization artifacts and research harnesses.}
Recent LLM-based methods for combinatorial optimization generate artifacts on the instance-solving path---heuristics, solving programs, and algorithmic procedures. ReEvo formulates LLMs as hyper-heuristics that iteratively improve heuristic designs \citep{ye2024reevo}. HeurAgenix couples offline heuristic evolution with state-aware online selection \citep{yang2025heuragenix}. TIDE separates structural generation from parameter tuning  \citep{chen2026tide}. DRoC retrieves decomposed constraint-relevant knowledge to support solving-program generation for vehicle routing \citep{jiang2025droc}, while ARS generates constraint-aware heuristic code from problem descriptions and retrieved constraints \citep{li2025ars}. More broadly, AlphaEvolve shows that LLMs can support discovery and improvement of algorithmic procedures traditionally requiring expert design \citep{novikov2025alphaevolve}. Other work uses LLMs to configure or control existing components, including online metaheuristic adaptation and cold-start separator configuration \citep{lawless2024coldstartseparator,xu2025autoep}. The closest line of work generates artifacts that operate inside solver systems. In the SAT setting, AutoSAT uses LLMs to optimize heuristic functions within SAT solvers \citep{sun2024autosat}, and AutoModSAT extends this to automatic discovery of code blocks and heuristics \citep{sun2025automodsat}. EvoCut uses evolution-guided LLMs to generate cuts for integer programming \citep{yazdani2025evocut}.

\paragraph{Harness engineering and in-context learning.}
Recent work identifies \emph{harness engineering}---the design of the orchestration layer governing information flow to the model---as a primary driver of agent performance \citep{anthropic2025harness,langchain2026anatomy,openai2026harness}. Meta-Harness shows that automated search over harness code can outperform hand-engineered baselines \citep{lee2026metaharness}; other work externalizes harness logic as natural-language artifacts \citep{pan2026nlah} or automatically synthesizes code harnesses \citep{lou2026autoharness}. Complementary agentic perspectives study long-running autonomous research workflows and multi-agent taxonomies \citep{haase2025beyondstatic,zimmer2026agenticresearcher}.
Self-Debugging shows that language models can improve generated programs by conditioning on execution results and debugging feedback through few-shot prompting \citep{chen2024selfdebugging}, while reflexion and self-refine use textual feedback or self-feedback as test-time context for iterative improvement without updating model weights \citep{madaan2023selfrefine,shinn2023reflexion}.
Recent studies examine how models interpret execution traces, self-generated tests, and feedback signals for code repair \citep{chen2025revisitselfdebugging,dai2025feedbackeval,ni2024next}.


\begin{figure}[H]
  \centering
  \resizebox{\textwidth}{!}{
    \begin{tikzpicture}[
        node distance=1.6cm,
        box/.style={
            rectangle,
            rounded corners,
            draw=black,
            very thick,
            align=center,
            minimum width=2.7cm,
            minimum height=1.0cm,
            fill=gray!10
          },
        mainbox/.style={
            rectangle,
            rounded corners,
            draw=black,
            very thick,
            align=center,
            minimum width=3.4cm,
            minimum height=1.35cm,
            font=\bfseries,
            text width=3.0cm
          },
        smallbox/.style={
            rectangle,
            rounded corners,
            draw=black,
            thick,
            align=center,
            minimum width=2.8cm,
            minimum height=0.95cm,
            fill=gray!6,
            text width=2.5cm
          },
        arrow/.style={->, very thick},
        dashedarrow/.style={->, very thick, dashed}
      ]

      \node[box] (prompt) {Prompt\\Specification};

      \node[mainbox, fill=green!15, right=of prompt] (generation) {Plugin\\Generation};

      \node[mainbox, fill=orange!18, right=of generation] (icl) {In-Context\\Learning};

      \node[mainbox, fill=purple!15, right=of icl] (signals) {Large-Scale\\Evaluation};

      \draw[arrow] (prompt) -- (generation);
      \draw[arrow] (generation) -- (icl);
      \draw[arrow] (icl) -- (signals);

      \node[smallbox, below=0.85cm of generation] (verification) {Plugin Update};
      \node[smallbox, below=0.85cm of icl] (diagnostics) {Feedback Signals};

      \draw[arrow] (icl) -- (diagnostics);
      \draw[arrow] (diagnostics) -- (verification);
      \draw[arrow] (verification) -- (generation);

      \begin{scope}[on background layer]
        \node[
          rectangle,
          rounded corners,
          draw=blue!80,
          very thick,
          dashed,
          fill=blue!5,
          inner sep=5pt,
          fit=(generation) (icl) (verification) (diagnostics),
          label={[font=\bfseries, text=blue!80]above:Harness Orchestration}
        ] (harness_box) {};
      \end{scope}

    \end{tikzpicture}
  }
  \caption{
    Overview of the Agentic MIPR Framework.
  }
  \label{fig:harness-pipeline}
\end{figure}
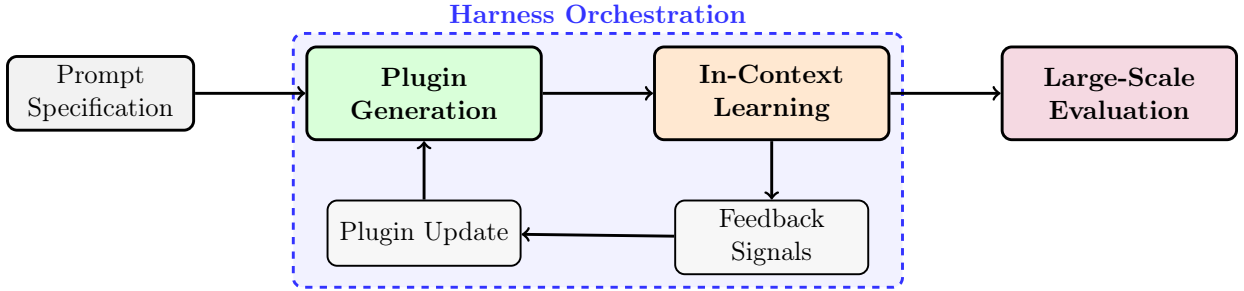

\section{A General Agentic MIPR Framework}
\label{sec:method}

A solver researcher typically starts from a hypothesis about formulation structure, propagation,
separation, branching, or primal search, and wants to test that hypothesis inside a branch-and-cut solver. Turning such a hypothesis into a working solver-side prototype requires
implementation, debugging, tuning, and benchmark validation---a process that is engineering-
intensive and tightly coupled to the solver core. The agentic MIPR framework targets exactly
this gap. Given a prompt-level specification of a solver component, it returns an
executable plugin that compiles, satisfies controlled correctness gates, and can be
benchmarked in a real-world setting. In particular, the framework can employ an agentic
researcher to formulate hypotheses from experiments and provide evaluative feedback. We
instantiate the framework in SCIP through PySCIPOpt~\citep{maher2016pyscipopt}, because SCIP
exposes a uniform plugin architecture. This paper focuses on one plugin
family: propagation-only constraint handlers, while the same harness design suggests
future extensions to other plugin families exposed by SCIP's broader plugin architecture.

\subsection{Framework Pipeline}

The framework drives LLM-based agents (e.g., Codex CLI, Claude Code, Gemini CLI) through three skill-controlled stages
(Figure~\ref{fig:harness-pipeline}):
\begin{enumerate}
  \item \textbf{Plugin Generation:} A user prompt is grounded against the knowledge base. The agent authors the mathematical core using pre-generated,
        interface-compliant templates, abstracting away boiler-plate solver logic.
  \item \textbf{In-context Learning:} The candidate plugin is executed on a targeted benchmark in a sandboxed environment. SCIP logs, error codes, and statistics are
        returned to the agent, which iteratively debugs and tunes the plugin using multi-objective feedback signals (soundness, time, node count, observed plugin handlers).  For example, soundness can be rejected when the default solver finds a feasible solution and  the generated plugin leads to infeasibility. This stage is also
        extended to handle exploration tasks, such as algorithmic search and plugin
        discovery.
  \item \textbf{Large-Scale Evaluation:} Validated plugins are dispatched for cluster-scale benchmarking (e.g., MIPLIB 2017). The agent synthesizes
        the logs and statistics to separate true algorithmic improvements from dominated, costly, or false-positives.
\end{enumerate}

\subsection{Harness and Resources for Structured Generation}
\label{subsec:harness-compact}

Because a generic code-generation loop cannot navigate tightly coupled branch-and-cut processes, the workflow is orchestrated by a solver-aware
\emph{harness}. Acting as a stateful wrapper, the harness mediates agent-solver interactions and persists artifacts across stages. It relies on three
structural elements:
\begin{enumerate}
  \item \textbf{Knowledge base:} A searchable repository of SCIP APIs, source-code excerpts, and mathematical references, served on-demand rather than
        via mega-prompts.
  \item \textbf{Templates:} Pre-generated interface-compliant scaffolds that handle common requirements like plugin registration, default parameters/methods, and state management. This
        eliminates SCIP API binding errors, allowing the agent to focus exclusively on component-specific logic.
  \item \textbf{Stage contracts:} Rigid specifications defining the inputs, outputs, and admissibility conditions for each stage. Bundled with
        references as agent skills, they create a stable boundary between human design and LLM generation.
\end{enumerate}
Across these stages, the agent and solver communicate via typed data records with harness-defined schemas, ensuring that generated components can be
tested and substituted independently within the constraint-handler generation workflow.

\subsection{In-context Learning in Sandbox}
\label{subsec:verification-compact}

The sandbox provides an isolated, controlled environment with access to real-world instances, such as those from the MIPLIB 2017 benchmark. Within this space, agents can execute generated plugins on these instances and receive comprehensive feedback, including solver logs, solving metrics, and detailed statistics on the plugins' behavioral execution.

In-context learning within the sandbox operates in two distinct operational modes: \emph{exploitation} and \emph{exploration}. In exploitation mode, a candidate plugin has already been generated. The agent acts as a coding agent and focuses on debugging and improving this plugin over a relatively small, specific dataset, such as instances where the target mathematical structures were previously detected. To evaluate the plugin's impact and correctness, the agent concurrently runs the default solver and directly compares the performance metrics.

In exploration mode, the agent acts as a researcher and explores a substantially larger dataset, potentially sketched  from the entire MIPLIB 2017 benchmark, to generate candidate plugin hypotheses.
This contrasts with the previous plugin generation stages, which rely on strong, highly
specified prompts accompanied by rich resources. Instead, exploration mode issues weak, open-ended prompts, such as ``discover a plugin that improves performance on logistics problems
.'' The exploration agent follows the generation pipeline specified by the harness, so the
templates and stage contracts are reused. A novelty gate compares candidate ideas against
existing SCIP handlers and already implemented project handlers, labels them as duplicates,
extensions, or novel candidates, and rejects candidates that merely reproduce known
propagation mechanisms under new names.

\subsection{Roles of the Agent}
\label{subsec:agent-roles-compact}

Agents in the framework take on distinct roles: an \emph{author} generating plugin code, a \emph{debugger} fixing  errors via execution
traces, a \emph{tuner} optimizing performance using in-context learning feedback, a \emph{researcher} discovering new constraint patterns, and an \emph{analyst}
processing benchmark results to guide future exploration. These roles share the same execution feedback loop, but operate at different levels of abstraction, from local code repair to benchmark-level hypothesis selection.

\section{Constraint Detection, Exploitation, and Exploration}
\label{sec:framework-scip}

As discussed in the introduction, domain propagation is a useful form of primal reduction and can improve MIP solving. In SCIP,  a propagation-only constraint handler applies semantics-aware domain propagation during node processing to tighten bounds before or during search.

We instantiate the general agentic MIPR framework for propagation-only constraint handlers and detectors.
A detector lifts a subset of linear rows of MIP problems into a semantic global constraint record; a generated constraint handler
registers this record with SCIP; and a propagator applies the corresponding bound-tightening rule.
For each constraint family \texttt{xyz}, the harness materializes code artifacts: a constraint detector \texttt{detector\_xyz.py}, a semantic data schema \texttt{constraint\_data\_xyz.py}, a SCIP constraint handler \texttt{conshdlr\_xyz.py}, and a propagator \texttt{propagator\_xyz.py}.

We next describe the additional technical details beyond the design of the framework in Section \ref{sec:method}.  For few-shot generation, detectors extract sets of linear constraints and encode them as CP global constraints; for zero-shot generation, they discover novel global constraint patterns from data. The resulting semantic representations are then used to generate dedicated constraint handlers.

\subsection{Prompt for Few-Shot Generation}
\label{subsec:prompt-skills}

We illustrate the prompt design for few-shot generation of a representative set of CP global constraints with well-documented semantics, propagation algorithms, and standard MIP encodings. Table~\ref{tab:constraint-families} summarizes this target space.

The few-shot generation task parallels traditional MIP modeling tasks, elevating the focus from translating natural-language problem specifications
into modeling languages \cite{ahmaditeshnizi2024optimus}, to translating plugin specifications directly into executable solver code.
For each global constraint, we design constraint-specific prompts
that exploit the structural correspondence in~\eqref{eq:global-linear-map}. These prompts provide the formal definition of the target constraint, specify the variable $(\mathbf{x},\mathbf{y})$ and parameter $c$ conventions, and define the expected structured output format. Specifically, we prompt agents to formalize both the forward encoding mapping ($\longrightarrow$) and the backward decoding mapping ($\longleftarrow$) from~\eqref{eq:global-linear-map}, which together specify the CP and MIP formulations required for detector generation. The prompts are further augmented with references to relevant academic literature and solver documentation on propagation methods. By packaging these references as progressively exposed agent skills, the user-level instruction is reduced to a compact directive, such as: ``Generate the detector, data schema, constraint handler, and propagator for the global constraint family \texttt{xyz} using the agent skill \texttt{cons\_xyz}.''
\begin{table}[htbp]
  \centering
  \caption{Existing global-constraint families and common MIP formulation patterns.}
  \label{tab:constraint-families}
  \small
  \begin{tabular}{@{}ll@{}}
    \toprule
    \textbf{Constraint Family} & \textbf{Common MIP Encoding}                                   \\
    \midrule
    \textsc{AllDifferent}      & Assignment matrices, pairwise disequality formulations         \\
    \textsc{Cardinality}       & Binary counting variables with linear cardinality bounds       \\
    \textsc{Channel}           & Linking equalities or implications between variable views      \\
    \textsc{Cumulative}        & Time-indexed formulations, disjunctive ordering variables      \\
    \textsc{NValue}            & Value-indicator variables with counting constraints            \\
    \textsc{Stretch}           & Transition variables, run-length indicators, local consistency \\
    \bottomrule
  \end{tabular}
\end{table}

\subsection{Synthetic Verification Harness}
\label{subsec:verification-layers}

The generated plugins are ensembles of code artifacts rather than trusted executable programs. An additional lightweight synthetic verification stage is added between the plugin generation and the in-context learning stages in Figure~\ref{fig:harness-pipeline}.
Rather than relying on external benchmarks, this procedure validates the plugins using small-scale, synthetic MIP instances and provides quick feedback.  Because the encoding formulation is already specified in the preceding stage,
the framework leverages reverse-sampling to construct synthetic MIP instances containing known global constraints $\text{Global Constraint}(x;c)$. To simulate the structural obfuscation typical of real-world models, we introduce random noise into these
instances: auxiliary linear constraints are injected, variables and rows are randomly permuted, and constraint signs are occasionally inverted:
\begin{equation}
  \label{eq:pipeline}
  \text{Global Constraint}(x;c)
  \;\xrightarrow{\text{reverse-sample}}\;
  \text{Instances}
  \;\xrightarrow{\text{detect}}\;
  \text{Global Constraint}(x';c').
\end{equation}

The first step verifies the soundness of the generated detector. As a semantic decoder, the detector is then tasked with parsing these obfuscated MIP instances to extract the embedded structure, outputting
$\text{Global Constraint}(x';c')$. The detector passes verification if and only if it exactly recovers the original structure and metadata (i.e., $x=x'$
and $c=c'$).

The second step verifies the validity of the propagator. We propose a general, enumeration-based validation method applicable to any primal reduction technique. Given a synthetic MIP instance with a detected global constraint, the agent leverages the known constraint specification to enumerate all feasible solutions, thereby explicitly constructing the true feasible region $X(c)$;  the generated propagator is then invoked to reduce this domain from a relaxed domain $D(c)$ to $D'(c)$:
\[
  \text{MIP instances}
  \;\xrightarrow{\text{enumerate}}\;
  X(c),
  \quad
  D(c)
  \;\xrightarrow{\text{propagate}}\;
  D'(c).
\]
Finally, the agent verifies the validity of the propagation via the inclusion check \eqref{eq:valid-reduction}.

\subsection{Exploitation and Exploration through In-Context Learning}

Within the sandbox environment, the agent tracks contextual feedback signals, including the number of detected constraints, the frequency of applied propagations, propagation time, and the total number of domain reductions induced. These quantities, which can be extracted from solver logs and statistics, are used to guide debugging and tuning. In addition, the global performance feedback is also provided, such as the number of solved instances within the time limit, overall running time, node counts, and duality gaps.

In exploitation mode, the agent works on a  generated constraint handler. The agent's tasks are to (i) repair any detection mismatches, (ii) analyze the feedback signals to iteratively modify the  handler, and (iii) optimize the  handler's performance.

In exploration mode, unlike the heavily structured prompts and skills for few-shot generation, the agent is given a simple open-ended prompt with an additional agent skill to explore the search space. We call this zero-shot generation. In particular, this can generate non-CP global constraints that include continuous variables. The exploration mode encourages the agent to deviate from its current strategy and search for novel opportunities using performance signals, while the novelty gate filters candidates that duplicate SCIP's existing handlers or previously generated project handlers.

\section{Experiments}
\label{sec:experiments}
The agentic MIPR harness is implemented using CLI-based coding agents augmented with skills, references, and code-access tools. All experiments are conducted using PySCIPOpt 6.0.0 with SCIP 10.0.0 and SoPlex 8.0.2. Experiments are performed on Intel Xeon Gold 5122 processors running at 3.60 GHz and 96 GB of RAM. All runs are conducted on a single thread with a time limit of two hours. We use the MIPLIB 2017 benchmark set of 1065 instances~\citep{mip2017}.

\subsection{Zero-Shot Generation Results}
\label{subsec:prompt-to-plugin-verification}

We report the results of zero-shot constraint-handler generation. In contrast to the per-family
few-shot prompts,  since
constraint patterns are discovered directly from MIP rows, they may include continuous variables and
multi-row structures not covered by a single well-known CP global constraint.

\begin{table}[htbp]
  \centering
  \scriptsize
  \setlength{\tabcolsep}{3pt}
  \renewcommand{\arraystretch}{1.08}
  \caption{Constraint patterns discovered via zero-shot generation.}
  \label{tab:novel_patterns}

  \begin{tabularx}{\linewidth}{@{}>{\raggedright\arraybackslash}p{0.24\linewidth}
    >{\raggedright\arraybackslash}X
    >{\raggedright\arraybackslash}X@{}}
    \toprule
    Name & Patterns                                                               & Propagation idea \\
    \midrule

    \textsc{OneHotResource}
         & One-hot choices coupled by a shared resource limit
         & Reason over remaining resource capacity to rule out infeasible choices                    \\

    \textsc{BottleneckExactOne}
         & Exact-one selector groups linked through a bottleneck variable
         & Propagate between selector feasibility and the bottleneck bound                           \\

    \textsc{RosteringWindow}
         & Shift assignment, coverage, and local work/rest-window rules
         & Use row-level residual reasoning to detect infeasible assignments                         \\

    \textsc{UnitCommitmentRamp}
         & Unit status, generation level, and ramping constraints over time
         & Propagate bounds across coupled status--power--ramp rows                                  \\

    \textsc{DisjPolyhedral}
         & Small big-M-encoded polyhedral disjunctions
         & Test feasible branches and combine their implied bound envelopes                          \\

    \bottomrule
  \end{tabularx}
\end{table}

Table~\ref{tab:novel_patterns} lists the five constraint handlers that survive
the hard novelty gate.  The constraint patterns can  capture multi-row mixed-integer
structures in real-world MIP instances that are not currently exploited by
the linear constraint handler in SCIP, which is designed to operate on
individual rows. These patterns may arise from MILP modeling of production planning, scheduling, and logistics problems, and include structures such as disjunctive constraints and logical relationships. The generated propagators then translate these recovered multi-row structures into local bound-tightening rules evaluated by the same benchmark harness.

\subsection{Computational Results and Analysis}
\label{subsec:propagation-configuration-ablation}

\paragraph{Configurations and baseline.} The baseline is the default SCIP configuration. We evaluate the impact of generated constraint handlers by
comparing the baseline against a \emph{plugin configuration}, where customized constraint handlers are explicitly activated at the root node. Since plugins are implemented as PySCIPOpt modules, the performance
might be weaken by python overhead and the lack of C-level efficiency.

\paragraph{Performance metrics.} We evaluate performance using standard MIP metrics~\citep{hojny2025scip10}: the number of solved instances within the time limit, shifted running times, and shifted node counts over solved instances. The number of solved instances filters out false positives that do not translate into practical gains, while running time captures overall performance and node count reflects search efficiency and pruning strength. To aggregate results, we use shifted geometric means, which are robust to outliers. The shifted geometric mean of values $x_i$ is
$
  \left(\prod_{i=1}^n (x_i + s)\right)^{1/n} - s.
$
Following SCIP conventions, we use $s=1$ for running time and $s=100$ for node counts.

\paragraph{Result organization.} Table~\ref{tab:pf-optimal-coverage} summarizes solving coverage, detailing the number of instances where constraints are \emph{Detected}, solved by both configurations
(\emph{Common}), or uniquely solved by the \emph{Baseline} or \emph{Plugin}. Table~\ref{tab:gmeans} evaluates performance on \emph{Common} instances solved by both, reporting geometric means for solving time (T) and node counts (N). Speedups are calculated as
baseline/plugin, where values $>1$ favor the plugin. We also report the number of instances where the plugin strictly improves time (\#T\textsubscript{speedup}) and node counts
(\#N\textsubscript{speedup}). In both tables, results for the six known constraint families are grouped at the top, while the five novel constraint patterns are at the bottom.

\begin{table}[htbp]
  \centering
  \scriptsize
  \setlength{\tabcolsep}{3pt}
  \renewcommand{\arraystretch}{1.05}
  \caption{Detected instances and solved instances by different configurations.}
  \label{tab:pf-optimal-coverage}
  \begin{tabular}{lrrrrrr}
    \toprule
    Family                      & Detected & Baseline & Plugin & Common & Baseline only & Plugin only \\
    \midrule
    \textsc{AllDifferent}       & 7        & 4        & 4      & 4      & 0             & 0           \\
    \textsc{Cardinality}        & 45       & 10       & 10     & 10     & 0             & 0           \\
    \textsc{Channel}            & 40       & 22       & 22     & 22     & 0             & 0           \\
    \textsc{Cumulative}         & 124      & 12       & 12     & 12     & 0             & 0           \\
    \textsc{NValue}             & 5        & 1        & 1      & 1      & 0             & 0           \\
    \textsc{Stretch}            & 1        & 1        & 1      & 1      & 0             & 0           \\
    \midrule
    \textsc{OneHotResource}     & 18       & 6        & 6      & 6      & 0             & 0           \\
    \textsc{BottleneckExactOne} & 4        & 0        & 2      & 0      & 0             & 2           \\
    \textsc{RosteringWindow}    & 8        & 4        & 6      & 4      & 0             & 2           \\
    \textsc{UnitCommitmentRamp} & 2        & 0        & 0      & 0      & 0             & 0           \\
    \textsc{DisjPolyhedral}     & 97       & 30       & 31     & 30     & 0             & 1           \\
    \bottomrule
  \end{tabular}
\end{table}

\begin{table}[htbp]
  \centering
  \tiny
  \setlength{\tabcolsep}{1pt}
  \renewcommand{\arraystretch}{1.05}
  \caption{Performance comparison on commonly solved optimal instances.}
  \label{tab:gmeans}
  \begin{tabular}{lrrrrrrrrr}
    \toprule
    Family                      & Common & T$_{\text{base}}$ & T$_{\text{plug}}$ & N$_{\text{base}}$ & N$_{\text{plug}}$ & T speedup & N speedup & \#T$_{\text{speedup}}$ & \#N$_{\text{speedup}}$ \\
    \midrule
    \textsc{AllDifferent}       & 4      & 97.15             & 108.38            & 458.27            & 458.27            & 0.896     & 1.000     & 2                      & 0
    \\
    \textsc{Cardinality}        & 10     & 39.24             & 40.43             & 485.16            & 485.16            & 0.971     & 1.000     & 3                      & 0
    \\
    \textsc{Channel}            & 22     & 215.74            & 208.84            & 1787.26           & 1799.94           & 1.033     & 0.993     & 11                     & 5
    \\
    \textsc{Cumulative}         & 12     & 130.49            & 131.81            & 2974.88           & 2974.88           & 0.990     & 1.000     & 6                      & 0
    \\
    \textsc{NValue}             & 1      & 20.09             & 20.29             & 1.00              & 1.00              & 0.990     & 1.000     & 0                      & 0
    \\
    \textsc{Stretch}            & 1      & 5.86              & 5.77              & 1.00              & 1.00              & 1.016     & 1.000     & 1                      & 0
    \\
    \midrule
    \textsc{OneHotResource}     & 6      & 167.54            & 375.22            & 1668.41           & 1668.41           & 0.447     & 1.000     & 1                      & 0
    \\
    \textsc{BottleneckExactOne} & 0      & --                & --                & --                & --                & --        & --        & 0                      & 0
    \\
    \textsc{RosteringWindow}    & 4      & 143.42            & 148.93            & 160.04            & 160.04            & 0.963     & 1.000     & 1                      & 0
    \\
    \textsc{UnitCommitmentRamp} & 0      & --                & --                & --                & --                & --        & --        & 0                      & 0
    \\
    \textsc{DisjPolyhedral}     & 30     & 193.69            & 192.58            & 2663.27           & 2431.83           & 1.006     & 1.095     & 10                     & 7
    \\
    \bottomrule
  \end{tabular}
\end{table}

\paragraph{Limitation of existing CP methods.} Regarding Q\ref{q:detection}, Table~\ref{tab:pf-optimal-coverage} confirms that MIPLIB 2017 contains rich CP global constraint structures.  While most few-shot CP handlers showed minimal impact in Table~\ref{tab:gmeans}, \textsc{Channel} yielded a modest 3.3\% time speedup alongside additional nodes. This
suggests that exploiting standard CP constraints in MIP may not yield significant benefits without further tuning or native C implementations.

\paragraph{Advantage of zero-shot exploration.} In contrast, the zero-shot
exploration successfully discovered  unexploited constraint patterns whose propagators solved five additional instances. This is a substantial signal, as SCIP 10.0 only
solves two more instances than version 9.0. Notably, \textsc{DisjPolyhedral} achieved time and node speedups of 0.6\% and 9.5\%, respectively, across 30 commonly solved instances.
This provides a positive answer to Q\ref{q:reduction}.

\paragraph{Additional ablation studies and observations.}
Supplementary diagnostics in Appendix~\ref{app:experimental-supplement} further explain the mixed performance of the generated handlers. A plugin generation reliability ablation (Appendix~\ref{app:artifact-reliability}) confirms that the full
  harness generates correct plugins, whereas direct unguided generation consistently fails. For the \textsc{Channel} handler, running the propagator at every node yields more cutoffs but often incurs enough overhead to offset the gains relative to root-only execution (Appendix~\ref{app:propfreq-comparison}). Finally, diagnostic audits (Appendix~\ref{app:cp-propagation-failure}) show that \textsc{AllDifferent} and \textsc{Cumulative} often yield neutral results because, despite correct detection and triggering, they produce no additional domain reductions beyond SCIP's default presolve and row-level reasoning.
\section{Conclusion}
\label{sec:conclusion}

We have presented an agentic MIP research (agentic MIPR) framework designed to accelerate research and solver development by fundamentally reducing the engineering
friction between algorithmic hypothesis and executable plugin in SCIP. Our primary contribution is a robust, solver-aware harness that integrates structured LLM
generation with rigorous verification mechanisms, such as reverse-sampling-based detector and propagation soundness checks and enumeration-based propagation soundness, coupled with sandbox-based repair and benchmark-scale evaluation. Applying this framework, we demonstrated its ability not only to recover
established global constraints from CP but also to systematically evaluate their impact on MIP solving.

Tuning and customizing general-purpose MIP solvers is notoriously difficult. Our experiments suggest that mining structures or patterns in MIP problems
is evolving toward  instance-specific approaches, where the activation of an algorithmic plugin depends on detecting patterns unique to specific
instances. This aligns with recent trends in instance-specific methodologies, such as AlphaEvolve \cite{novikov2025alphaevolve}. In this context, the
agentic in-context learning paradigm offers a distinct advantage over traditional supervised learning: it is computationally lightweight and requires no
updates to model weights. More importantly, it enables the discovery of genuinely novel, MIP-native semantic patterns via zero-shot exploration. The
core novelty of this approach lies in autonomously translating problem structures and experimental feedback directly into formalized solver logic.

We anticipate a bright future for agentic MIPR, with natural extensions to other plugin families and more sophisticated exploration strategies. By bridging the gap between human insight and solver implementation, we hope to democratize and accelerate MIP research through new
optimization interfaces.

\bibliographystyle{plainnat}
\bibliography{references}


\appendix

\section{Prompt, Constraint Patterns, and Propagation Logic in Zero-Shot Generation}
\label{section:zero-shot-details}

This appendix gives the formal definitions of the five zero-shot constraint
patterns of Table~\ref{tab:novel_patterns} and a worked
prompt-to-artifact example for the generated zero-shot handlers. The exploration prompt under which
the agent operates and the hard novelty gate that filters surviving
candidates are stated in Section~\ref{subsec:prompt-to-plugin-verification};
we focus here on what those candidates contain and on the inference each
generated handler performs.

\FloatBarrier
\subsection{Zero-Shot Constraint Pattern Definitions}
\label{app:zero-shot-pattern-definitions}

For each pattern we give the abstracted MIP row block and the propagation
rule that distinguishes the generated handler from existing SCIP propagation. For zero-shot generation, we use the general exploration prompt template, augmented with a few structure- and application-specific sentences. We detail these
additional sentences below for each generated constraint handler.
We write \emph{residual-activity tightening} for the standard rule that, on a
row $\ell \le \sum_i a_i x_i \le r$, tightens
$x_j \le (r - \mathrm{minact}_{-j})/a_j$ and
$x_j \ge (\ell - \mathrm{maxact}_{-j})/a_j$ for $a_j>0$, with sign-reversed
updates for $a_j<0$. Unlike SCIP's linear handler, which applies this rule
row-by-row in a global pool, the generated handlers run it inside the detected
\emph{semantic multiple rows}.

\begin{example}[\textsc{OneHotResource}]
  \noindent\textbf{Additional prompt sentence:}
  \emph{None.}

  Disjoint binary one-hot groups share one capacity row,
  \[
    \sum_{k} y_{g,k} = 1 \;\; \forall g,
    \qquad
    r_{\min} + \sum_{g} \sum_{k} c_{g,k}\, y_{g,k} \le B,
  \]
  where $r_{\min}$ is the best-case capacity contribution of variables
  outside the one-hot groups under their initial bounds. Let
  $m_g = \min\{c_{g,k}:\operatorname{ub}(y_{g,k})>0\}$ and
  $M = r_{\min} + \sum_g m_g$. The propagator returns cutoff if $M>B$, and
  otherwise removes any option that cannot fit while every \emph{other}
  group takes its cheapest remaining choice,
  \[
    c_{g,k} > B - (M - m_g)
    \;\;\Longrightarrow\;\;
    \operatorname{ub}(y_{g,k}) \leftarrow 0,
  \]
  iterating with standard one-hot fix-point closure. The novelty over plain
  \texttt{setppc} plus a single-row knapsack handler is the cross-group
  residual reasoning that uses the cheapest alternative in every other group.
\end{example}

\begin{example}[\textsc{BottleneckExactOne}]
  \noindent\textbf{Additional prompt sentence:}
  \emph{Consider problems and structures with logical relationships.}

  Exact-one selector groups are linked to a shared continuous bottleneck
  variable~$z$,
  \[
    \sum_{j} x_{i,j} = 1 \;\; \forall i,
    \qquad
    z \ge w_{i,j}\, x_{i,j} \;\; \forall i,j .
  \]
  The propagator pushes
  $\operatorname{lb}(z) \leftarrow
    \max_{i} \min\{w_{i,j} : \operatorname{ub}(x_{i,j}) > 0\}$ and removes
  selectors above the current upper bound,
  $w_{i,j} > \operatorname{ub}(z) \Rightarrow \operatorname{ub}(x_{i,j})
    \leftarrow 0$. An activation-aware variant adds binary activators $y_j$
  with $x_{i,j}\le y_j$ and an exact-cardinality row $\sum_j y_j = p$, on
  which the propagator additionally enforces the implications
  $x_{i,j}=1\Rightarrow y_j=1$, $y_j=0\Rightarrow x_{i,j}=0$, and a
  coverage-based radius rule: for a candidate radius~$R$, if a
  disjoint-neighbourhood packing of the activators reachable through
  selectors with $w_{i,j}<R$ requires more than~$p$ activators, then
  $\operatorname{lb}(z) \leftarrow R$ (located by binary search). The novelty
  over per-row generic linear propagation is the cross-group $\max\!\min$
  inference between selector domains and~$z$.
\end{example}

\begin{example}[\textsc{RosteringWindow}]
  \noindent\textbf{Additional prompt sentence:}
  \emph{Consider problems and structures in production planning. Propose up to five prototype handlers passing novelty gates.}

  An integrated nurse-rostering block over binary assignments
  $x_{n,d,s} \in \{0,1\}$,
  \[
    \sum_{s} x_{n,d,s} \le 1,
    \qquad
    \sum_{n} x_{n,d,s} = \mathrm{req}_{d,s},
  \]
  together with window, weekend, and shift-registration row families
  (\texttt{workWind}, \texttt{restWind}, \texttt{flow}, \texttt{dem},
  \texttt{sos}, \texttt{hlb}, \texttt{hub}, \texttt{lba}, $\dots$) detected
  by row-family patterns and row-variable incidence. The propagator iterates
  residual-activity tightening across the full block. The novelty is the
  block detection itself: SCIP propagates these rows in a global pool, while
  the generated handler keeps them as one coupled unit and runs a localised
  fix-point that can discover reductions a single-pass linear handler misses.
\end{example}

\begin{example}[\textsc{UnitCommitmentRamp}]
  \noindent\textbf{Additional prompt sentence:}
  \emph{Consider problems and structures in production planning.  Propose up to five prototype handlers passing novelty gates.}

  A unit-commitment block coupling, for each generator~$g$ and period~$t$,
  binary status $u_{g,t}$, startup $v_{g,t}$, and shutdown $w_{g,t}$ with
  continuous power $p_{g,t}$ and reserve $r_{g,t}$,
  \[
    u_{g,t} - u_{g,t-1} = v_{g,t} - w_{g,t},
    \qquad
    P^{\min}_g\, u_{g,t} \le p_{g,t} \le P^{\max}_g\, u_{g,t},
  \]
  \[
    \sum_g p_{g,t} \ge D_t,
    \quad
    \sum_g r_{g,t} \ge R_t,
    \quad
    p_{g,t} - p_{g,t-1} \le RU_g\, u_{g,t-1} + SU_g\, v_{g,t},
  \]
  with the symmetric ramp-down row, detected by stratifying selected rows by
  role (\texttt{Demand}, \texttt{Reserve}, \texttt{Link}, \texttt{Ramp},
  \texttt{MinUp}, $\dots$) so that late-occurring roles are not dropped by
  row caps. As in \textsc{RosteringWindow}, propagation is residual-activity
  tightening across the block; the novelty is again block detection rather
  than per-row inference.
\end{example}

\begin{example}[\textsc{DisjPolyhedral}]
  \noindent\textbf{Additional prompt sentence:}
  \emph{Consider structures of disjunctive constraints that are relaxed by big-M.}

  A small polyhedral disjunction encoded by big-M-guarded inequalities,
  \[
    \bigvee_{i=1}^{B} \bigl( A_i\, x \le b_i \bigr),
    \qquad B \le 6,
  \]
  detected in two variants: a binary-selector disjunction (rows sharing one
  guard~$y$ with both $y=0$ and $y=1$ branches populated) and an exact-one
  mode disjunction ($\sum_{i=1}^B y_i = 1$ with each $y_i$ activating one
  polytope). A row enters the disjunction only if its guard coefficient is
  large enough that the inactive side is dominated by the maximum activity
  of the other terms under the original bounds.

  The propagator implements a constructive-disjunction rule. For each
  disjunction it runs residual-activity tightening on every branch's active
  rows; marks a branch \emph{impossible} when its polytope becomes empty,
  removing the corresponding selector value; returns cutoff if no branch is
  viable; fixes the selector when only one branch survives; and otherwise
  tightens each touched variable to the envelope of the surviving
  branch-local bounds,
  \[
    \operatorname{lb}(x) \leftarrow
    \min_{i \in \text{viable}} \operatorname{lb}^{(i)}(x),
    \qquad
    \operatorname{ub}(x) \leftarrow
    \max_{i \in \text{viable}} \operatorname{ub}^{(i)}(x).
  \]
  This goes beyond the row-level big-M propagation that SCIP applies to the
  linearised indicator rows in isolation: SCIP's linear handler cannot take
  the union of viable branch domains because the rows live in a global pool
  with no disjunction structure.
\end{example}

\FloatBarrier
\subsection{\textsc{BottleneckExactOne} Worked Example}
\label{app:zero-shot-bottleneck-worked-example}

This example traces \textsc{BottleneckExactOne} from the open-ended
exploration prompt of Section~\ref{subsec:prompt-to-plugin-verification} to a
propagation-only SCIP constraint handler.

\paragraph{Exploration grounding.}
From assignment-style support rows linked to a shared continuous variable,
the agent isolates the max-style structure of
Appendix~\ref{app:zero-shot-pattern-definitions}: each demand group selects
exactly one alternative ($\sum_j x_{i,j}=1$) and a shared variable~$z$
upper-bounds the selected weight ($z \ge w_{i,j} x_{i,j}$). The novelty
gate accepts the candidate because \texttt{setppc} covers only the exact-one
rows and generic linear residual activity propagates each link row in
isolation, missing the cross-group $\max\!\min$ inference.

\paragraph{Constraint detection.}
The detector scans for exact-one rows over binary selectors, groups
bottleneck-link rows by the selector row and the shared bottleneck variable,
optionally records the activation rows $x_{i,j}\le y_j$ and $\sum_j y_j=p$,
and rejects matches whose evidence on any of the three components is too
weak. It emits a semantic record of selector groups, weights, the CP
position of~$z$, and optional activator metadata.

\paragraph{Artifact synthesis.}
The agent generates the runtime modules consumed by the SCIP registration path:
\begin{enumerate}
  \item \texttt{constraint\_data\_bottleneck\_exactone.py}: parsed semantic
        schema with fields \texttt{groups}, \texttt{weights},
        \texttt{bottleneck\_index}, and the optional \texttt{activator\_indices},
        \texttt{selector\_activator\_indices}, \texttt{open\_count}.
  \item \texttt{conshdlr\_bottleneck\_exactone.py}: SCIP-facing bridge.
        \texttt{createCons} attaches the serialized record;
        \texttt{consprop} reads local bounds, calls the propagator, and applies
        safe \texttt{tightenVarLb}/\texttt{tightenVarUb} updates while skipping
        inactive or multi-aggregated variables; the remaining callbacks stay
        passive feasible. The default registration uses \texttt{propfreq=0}
        after sandbox evidence indicated frequent Python-callback propagation
        was too expensive.
  \item \texttt{propagator\_bottleneck\_exactone.py}: implements the rule of
        Appendix~\ref{app:zero-shot-pattern-definitions}, iterating exact-one
        consequences, the bottleneck lower bound, the selector cut, the
        optional activator implications, residual exact-cardinality
        propagation, and the coverage-based radius rule until a fixed point.
\end{enumerate}

\begin{table}[H]
  \centering
  \caption{Artifact trace for the \textsc{BottleneckExactOne} worked example.}
  \label{tab:zero-shot-workflow-bottleneck}
  \small
  \begin{tabular}{@{}ll@{}}
    \toprule
    \textbf{Pipeline stage} & \textbf{Representative artifact or file}                                          \\
    \midrule
    Exploration prompt      & open-ended directive of Section~\ref{subsec:prompt-to-plugin-verification}        \\
    Skill                   & \texttt{exploration} skill with hard novelty gate                                 \\
    Detector                & \texttt{codes/detector\_bottleneck\_exactone.py}                                  \\
    Semantic record schema  & \texttt{codes/constraint\_data\_bottleneck\_exactone.py}                          \\
    SCIP integration layer  & \texttt{codes/conshdlr\_bottleneck\_exactone.py}                                  \\
    Propagation logic       & \texttt{codes/propagator\_bottleneck\_exactone.py}                                \\
    Verification workflow   & sandbox MIP instances, soundness gate, default-vs-handler propagation diagnostics \\
    \bottomrule
  \end{tabular}
\end{table}

\section{Experimental Supplement}
\label{app:experimental-supplement}
\label{app:signal-taxonomy}

\subsection{Artifact-Production Reliability Study}
\label{app:artifact-reliability}

This appendix-only diagnostic study evaluates the harness as an
artifact-production pipeline rather than as a solver-performance method.  We
compare direct LLM generation with the full harness on three core CP-family
targets for which the common detector/handler/propagator verification pipeline
applies: \textsc{Cardinality}, \textsc{Channel}, and \textsc{Cumulative}.
Direct LLM generation uses minimal project context and generates the detector,
constraint data, constraint handler, and propagator once, without
family-specific skills, verification feedback, or iterative repair.  The full
harness uses the existing skill/template/verification/repair pipeline with a
repair budget of three iterations per target.

Both settings are evaluated with the operational gates summarized in
Table~\ref{tab:artifact-reliability-gates}. The load gate uses a minimal
PySCIPOpt model. Detector verification uses the synthetic
\texttt{constraint\_CI} generator with \texttt{--num-runs 1},
\texttt{--seed 7}, and \texttt{--compare-level full}. Propagator soundness uses the synthetic propagation verifier with seed 11 and bounded enumeration. The smoke gate runs the plugin on the smallest available family-specific MPS
instance under \texttt{data/constraint/cons\_<family>/}. Since direct LLM artifacts fail detector verification, their smoke gate is not reached.
The evaluator subprocess timeouts are 30 seconds for load, 120 seconds for
detector verification, 120 seconds for propagator soundness, and 90 seconds for
smoke; the smoke solve uses a 5-second SCIP time limit.

\begin{table}[h]
  \centering
  \caption{Operational gates used in the artifact-production reliability study.}
  \label{tab:artifact-reliability-gates}
  \scriptsize
  \setlength{\tabcolsep}{3pt}
  \begin{tabular}{@{}p{0.22\linewidth}p{0.52\linewidth}p{0.18\linewidth}@{}}
    \toprule
    \textbf{Gate} & \textbf{Pass criterion} & \textbf{Limit} \\
    \midrule
    Artifact completeness &
    Required detector, constraint-data, handler, and propagator files are present. &
    -- \\
    Syntax/import &
    Generated Python files pass \texttt{py\_compile}. &
    -- \\
    SCIP/PySCIPOpt load &
    Modules import and register in a minimal PySCIPOpt model. &
    30 s \\
    Detector verification &
    Synthetic CI check reports no mismatches or errors. &
    120 s \\
    Propagator soundness &
    Bounded-enumeration soundness checker exits successfully. &
    120 s \\
    Benchmark smoke &
    Family-specific MPS smoke solve completes without error. &
    90 s; SCIP 5 s \\
    Benchmark-ready &
    All preceding gates pass. &
    -- \\
    \bottomrule
  \end{tabular}
\end{table}

Table~\ref{tab:artifact-reliability} summarizes the resulting
artifact-production funnel.  Direct LLM artifacts passed artifact completeness,
syntax/import, and SCIP/PySCIPOpt load for all three targets, but failed
detector verification in all three cases: during full-mode preflight, the
detectors returned empty record lists.  The full harness produced
benchmark-ready artifacts for all three core targets.  The \textsc{Cardinality}
run also records a conservative verification-guided repair.  Detector
verification exposed empty \textsc{Cardinality} records, after which a
synthetic-generator-compatible detector branch was added.  Propagation
soundness then reported a cutoff despite feasible assignments, so the
propagator was made conservative.

\begin{table}[h]
  \centering
  \caption{Artifact-production reliability on three core CP-family targets.}
  \label{tab:artifact-reliability}
  \scriptsize
  \setlength{\tabcolsep}{4pt}
  \begin{tabular}{@{}lrrrrrr@{}}
    \toprule
    \textbf{Setting} &
    \textbf{Targets} &
    \begin{tabular}[c]{@{}c@{}}\textbf{Syntax/}\\\textbf{load}\end{tabular} &
    \textbf{Detector} &
    \textbf{Propagator} &
    \textbf{Smoke} &
    \begin{tabular}[c]{@{}c@{}}\textbf{Benchmark}\\\textbf{ready}\end{tabular} \\
    \midrule
    Direct LLM   & 3 & 3 & 0 & 0 & 0 & 0 \\
    Full harness & 3 & 3 & 3 & 3 & 3 & 3 \\
    \bottomrule
  \end{tabular}
\end{table}

This study supports a narrow artifact-production claim: in this controlled
diagnostic setting, the full harness produced executable SCIP/PySCIPOpt
artifacts that passed the available local verification and benchmark-smoke
gates for the three core CP-family targets, while direct generation did not.
It also shows that compile/load success alone is not sufficient for
solver-artifact validity, since detector verification rejected syntactically
valid and loadable but semantically ineffective direct LLM artifacts.

This is a controlled diagnostic reliability study, not a solver-performance
benchmark.  It does not evaluate MIPLIB speedup, does not establish statistical
significance, and uses one run per family and setting.  It does not prove that
direct LLM generation is generally poor, that repaired artifacts are
propagation-strength optimal, or that generated plugins are globally correct or
production-ready.

\subsection{Propagation frequency comparison}
\label{app:propfreq-comparison}

To further analyze the impact of propagation frequency, we evaluate two
settings of the handler propagation-frequency parameter \texttt{propfreq} on
the Channel-detected instances.  The setting \texttt{propfreq=0} runs the generated
constraint handler and propagator only at the root node, while \texttt{propfreq=1}
enables propagation at every eligible branch-and-bound node.  We use a SCIP
time limit of 300 seconds.  Since
\texttt{propfreq=1} changes how much of the time budget is spent inside generated
callbacks, shifted running time and node count alone are difficult to interpret:
fewer processed nodes may reflect stronger pruning, but may also reflect that
the solver spent much of the budget inside propagation callbacks.  We therefore
use this comparison as a diagnostic study of propagation calls, domain
reductions, cutoffs, and propagation time.  The purpose is to explain the
formal benchmark behavior reported in the main text: \textsc{Channel} might be tuned for better performance, but even there the benefit depends on where and
how often the generated handler is invoked.

The per-run diagnostics in Table~\ref{tab:channel-propfreq-diagnostics}
report shifted geometric means for generated-handler calls, domain reductions,
cutoffs, and propagation time, using shift~$1$ to handle zero-valued diagnostic
counts; the zero-reduction column remains a run count.  These quantities are
not SCIP performance scores; they summarize typical plugin activity across
paired runs.  Moving from root-only to all-node
propagation increases the shifted geometric mean of generated-handler calls
from $8.6$ to $19{,}637.3$ and the shifted geometric mean of measured
\textsc{Channel} propagation time from $0.22$ seconds to $114.1$ seconds.  This
pattern is consistent with repeated Python-level constraint-handler callbacks
slowing down the search.  This helps explain why the formal experiments do not
support a blanket conclusion that more semantic propagation is better: even for
\textsc{Channel}, additional inference can be offset by the cost of invoking
the generated propagator too frequently.

\begin{table}[h]
  \centering
  \caption{\textsc{Channel} propagation diagnostics on paired valid instances.
    Diagnostic activity columns report shifted geometric means with shift~$1$,
    except for zero-reduction runs, which are counted.}
  \label{tab:channel-propfreq-diagnostics}
  \small
  \begin{tabular}{@{}lrrrrr@{}}
    \toprule
    \textbf{Setting}   & \textbf{Calls}          & \textbf{Domain red.}    &
    \textbf{Cutoffs}   & \textbf{Zero-red. runs} & \textbf{Prop. time (s)} \\
    \midrule
    \texttt{root-only} & 8.6                     & 5.0                     & 0.0 & 15 & 0.22  \\
    \texttt{all-node}  & 19{,}637.3              & 50.3                    & 8.6 & 18 & 114.1 \\
    \bottomrule
  \end{tabular}
\end{table}

In 18 paired runs, all-node propagation was executed but produced no domain
reductions.  This supports the interpretation that, on some instances, the
Channel inference available to the generated plugin is already weak or
redundant after SCIP's presolve and row-level reasoning.

The instance-level examples in
Table~\ref{tab:channel-root-reduction-allnode-zero-examples} focus on runs
where root-only propagation finds reductions, but all-node propagation later
triggers many times without new domain reductions.  Root propagation and
all-node propagation are therefore not simply ``less'' and ``more'' of the same
useful work.  The root node may still contain broad global structure, while
subnodes have already been shaped by branching, presolve, and row-level
propagation; repeated semantic checks at those subnodes may therefore have
little new information to exploit.

\begin{table}[h]
  \centering
  \caption{Raw instance-level \textsc{Channel} examples comparing root-only and
    all-node propagation activity.}
  \label{tab:channel-root-reduction-allnode-zero-examples}
  \scriptsize
  \setlength{\tabcolsep}{5pt}
  \renewcommand{\arraystretch}{1.05}
  \begin{tabular}{@{}lrrrr@{}}
    \toprule
    \textbf{Instance}                   &
    \multicolumn{2}{c}{\textbf{Root-only}} &
    \multicolumn{2}{c}{\textbf{All-node}} \\
    \cmidrule(lr){2-3}\cmidrule(l){4-5}
                                        & \textbf{Calls} & \textbf{DomRed} &
    \textbf{Calls}                      & \textbf{DomRed} \\
    \midrule
    \texttt{cvrpb-n45k5vrpi}            & 2   & 1   & 2{,}476  & 0 \\
    \texttt{mrcpspj30-15-5i}            & 175 & 37  & 48{,}216 & 0 \\
    \texttt{mrcpspj30-53-3i}            & 119 & 19  & 44{,}695 & 0 \\
    \texttt{gfd-schedulen55f2d50m30k3i} & 28  & 952 & 9{,}639  & 0 \\
    \bottomrule
  \end{tabular}
\end{table}

\subsection{Failure of CP Global-Constraint Propagation}
\label{app:cp-propagation-failure}

The agentic evaluation also exposes a more basic failure mode: many
prompt-level ideas pass recognition and integration checks, but still do not
create useful propagation.  We audit the generated \textsc{AllDifferent},
\textsc{Cardinality}, and \textsc{Cumulative} handlers on the
detected-and-registered instances, using a 100-second SCIP time limit.  This
short run is intended as a diagnostic audit, not as a final performance
benchmark.  The main experiments compare against SCIP's default baseline; here
we explain why these generated CP-propagation candidates can be classified as
failures or low-value positives in that broader evaluation.  In other words,
this appendix connects the benchmark-level observation---most CP families do
not improve over the SCIP baseline---to solver statistics showing that the
generated handlers often add little or no propagation.

The counts below only include runs in which the detector found the target
structure, the harness registered at least one generated constraint handler,
and SCIP statistics were written; runs that hit the external watchdog are
counted separately as errors.  The root-only mode is
\texttt{propfreq=0}, while the all-node mode is \texttt{propfreq=1}.

The activity counts in Table~\ref{tab:cp-propagation-activity} collects three
cases: registered handlers whose propagators never run, registered handlers
whose propagators run but produce zero domain reductions, and registered
handlers that produce at least one domain reduction in all three constraints types.
The generated handlers are often successfully registered, but propagation either does not run or runs
without producing domain reductions.  This is not necessarily a detector
failure: detection only says that a group of rows matches a semantic family; it
does not imply that the recovered semantic object has additional
bound-tightening power at the current SCIP domains.  The generated propagators
are also conservative, so a correct handler can return no change when no
strictly tighter bound can be proven.  Finally, SCIP's presolve and row-level
reasoning may already expose the easy implications, leaving little additional
work for the generated semantic propagator.  These cases are therefore
low-value positives found by the agentic evaluation: they are semantically
recognized and safely integrated, but they do not add useful propagation beyond
what the solver already obtains.  This is the mechanism behind many of the
negative or neutral outcomes in the formal benchmark table.

\begin{table}[h]
  \centering
  \caption{Propagation activity for generated CP global-constraint handlers.}
  \label{tab:cp-propagation-activity}
  \small
  \setlength{\tabcolsep}{3pt}
  \begin{tabular}{@{}llrrrrr@{}}
    \toprule
    \textbf{Family} & \textbf{Setting} & \textbf{Reg.} &
    \begin{tabular}[c]{@{}c@{}}\textbf{Zero}\\\textbf{prop.}\end{tabular} &
    \begin{tabular}[c]{@{}c@{}}\textbf{Prop., zero}\\\textbf{DomRed}\end{tabular} &
    \begin{tabular}[c]{@{}c@{}}\textbf{Positive}\\\textbf{DomRed}\end{tabular} &
    \begin{tabular}[c]{@{}c@{}}\textbf{Max prop.}\\\textbf{time (s)}\end{tabular} \\
    \midrule
    \textsc{AllDifferent}    & root-only                   & 36            & 1 & 34 & 1 & 16  \\
    \textsc{AllDifferent}    & all-node                    & 36            & 0 & 34 & 2 & 100 \\
    \textsc{Cardinality}     & root-only                   & 39            & 6 & 32 & 1 & 3   \\
    \textsc{Cardinality}     & all-node                    & 40            & 4 & 31 & 5 & 100 \\
    \textsc{Cumulative}      & root-only                   & 42            & 2 & 38 & 2 & 9   \\
    \textsc{Cumulative}      & all-node                    & 42            & 0 & 35 & 7 & 100 \\
    \bottomrule
  \end{tabular}
\end{table}

The examples in Table~\ref{tab:cp-heavy-positive-propagation} cover another
type of case: generated propagators that do produce domain reductions and
cutoffs under all-node propagation, but spend a large fraction of the
100-second limit doing so.  The table reports raw per-instance statistics,
because the purpose is to show concrete heavy-propagation cases rather than an
aggregate trend.  These cases are not semantic failures; rather, they are
low-value positives for the current SCIP/PySCIPOpt implementation.  They show
why benchmark-scale feedback is necessary: a generated plugin can be correct,
detect real structure, and make local inferences, while still being too
expensive to be useful.  Useful CP-style domain reduction likely needs stronger
filtering, cheaper triggering rules, or a lower-overhead implementation before
it can improve solver behavior reliably.

\begin{table}[h]
  \centering
  \caption{Raw all-node examples where generated propagation produces reductions
    but consumes substantial time.}
  \label{tab:cp-heavy-positive-propagation}
  \scriptsize
  \setlength{\tabcolsep}{3pt}
  \begin{tabular}{@{}llrrrrr@{}}
    \toprule
    \textbf{Family}       & \textbf{Instance}            & \textbf{Calls}          &
    \textbf{DomRed}       & \textbf{Cutoffs}             & \textbf{Prop. time (s)} &
    \textbf{Nodes}                                                                                                     \\
    \midrule
    \textsc{AllDifferent} & \texttt{fhnw-sq2}            & 7{,}403                 & 1{,}704 & 29      & 97 & 1{,}024  \\
    \textsc{Cardinality}  & \texttt{ponderthis0517-inf}  & 10{,}995                & 677     & 144     & 96 & 3{,}035  \\
    \textsc{Cumulative}   & \texttt{mad}                 & 13{,}426                & 40      & 10      & 80 & 2{,}768  \\
    \textsc{Cumulative}   & \texttt{graphdraw-gemcutter} & 48{,}271                & 2{,}580 & 37      & 75 & 8{,}299  \\
    \textsc{AllDifferent} & \texttt{l2p12}               & 8{,}791                 & 32      & 4       & 71 & 32       \\
    \textsc{Cardinality}  & \texttt{mad}                 & 43{,}436                & 6{,}032 & 1{,}183 & 70 & 16{,}996 \\
    \bottomrule
  \end{tabular}
\end{table}

\end{document}